\documentclass{article} 
\usepackage{iclr2024_conference,times}

\usepackage{amsmath,amsfonts,bm}

\def\eqref#1{equation~\ref{#1}}

\def\1{\bm{1}}

\DeclareMathAlphabet{\mathsfit}{\encodingdefault}{\sfdefault}{m}{sl}
\SetMathAlphabet{\mathsfit}{bold}{\encodingdefault}{\sfdefault}{bx}{n}

\usepackage[utf8]{inputenc} 
\usepackage[T1]{fontenc}    
\usepackage[colorlinks,
linkcolor=red,
anchorcolor=cyan,
citecolor=cyan,
]{hyperref}       
\usepackage{url}            
\usepackage{booktabs}       
\usepackage{amsfonts}       
\usepackage{nicefrac}       
\usepackage{microtype}      
\usepackage{xcolor}         

\usepackage{natbib}
\usepackage{multirow}
\usepackage{tcolorbox}
\usepackage{tikz}
\tcbuselibrary{skins}
\usepackage{pifont}
\usepackage{caption}
\usepackage{subfigure}
\usepackage[graphicx]{realboxes}
\usepackage{wrapfig}
\tcbuselibrary{raster}
\usepackage{tabularx}
\usepackage{amsmath}
\tcbuselibrary{breakable}

\definecolor{darkpastelgreen}{rgb}{0.01, 0.75, 0.24}
\NewDocumentCommand{\heng}
{ mO{} }{\textcolor{red}{\textsuperscript{\textit{Heng}}\textsf{\textbf{\small[#1]}}}}

\NewDocumentCommand{\ziqi}
{ mO{} }{\textcolor{blue}{\textsuperscript{\textit{Ziqi}}\textsf{\textbf{\small[#1]}}}}
\NewDocumentCommand{\chihan}
{ mO{} }{\textcolor{green}{\textsuperscript{\textit{Chi}}\textsf{\textbf{\small[#1]}}}}
\NewDocumentCommand{\pengfei}
{ mO{} }{\textcolor{violet}{\textsuperscript{\textit{Pengfei}}\textsf{\textbf{\small[#1]}}}}
\NewDocumentCommand{\zhenhailong}
{ mO{} }{\textcolor{cyan}{\textsuperscript{\textit{Zhenhailong}}\textsf{\textbf{\small[#1]}}}}

\NewDocumentCommand{\tianci}
{ mO{} }{\textcolor{blue}{\textsuperscript{\textit{tianci}}\textsf{\textbf{\small[#1]}}}}

\title{\ours{}: Detecting and Rectifying Factual Inconsistency  in Reasoning by Reversing Chain-of-Thought}

\author{%
  Tianci Xue$^1$\thanks{Work done during the internship at UIUC.}    , Ziqi Wang$^2$, Zhenhailong Wang$^2$, Chi Han$^2$, Pengfei Yu$^2$, Heng Ji$^2$ \\
  $^1$ Department of Software, Nanjing University \\
  $^2$ Department of Computer Science, University of Illinois Urbana-Champaign \\
  \texttt{xuetianci@smail.nju.edu.cn}\\
  \texttt{\{ziqiw9, wangz3, chihan3, pengfei4, hengji\}@illinois.edu} \\
}

\newcommand{\oursfull}{Reversing Chain-of-Thought}
\newcommand{\ours}{\textsc{RCoT}}
\newcommand{\probone}{hallucination}
\newcommand{\probtwo}{overlooking}
\newcommand{\probthree}{misinterpretation}
\newcommand{\voneh}{hallucinate}
\newcommand{\vtwo}{overlook}
\newcommand{\vthree}{misinterpret}

\newcommand{\edone}{hallucinated}
\newcommand{\edtwo}{overlooked}
\newcommand{\edthree}{misinterpreted}
\newcommand{\word}{revise}

\iclrfinalcopy 
\begin{document}

\maketitle

\begin{abstract}
  Large language Models (LLMs) have achieved promising performance on arithmetic reasoning tasks by incorporating step-by-step chain-of-thought (CoT) prompting. 
  However, LLMs face challenges in maintaining factual consistency during reasoning, exhibiting tendencies to condition overlooking, question misinterpretation, and condition hallucination over given problems.
  Existing methods use coarse-grained feedback (e.g., whether the answer is correct) to improve factual consistency. 
  In this work, we propose \ours{} (\textbf{R}eversing \textbf{C}hain-\textbf{o}f-\textbf{T}hought), a novel method to improve LLMs' reasoning abilities by automatically detecting and rectifying factual inconsistency in LLMs' generated solutions. 
  To detect factual inconsistency, \ours{} first asks LLMs to reconstruct the problem based on generated solutions. Then fine-grained comparisons between the original problem and the reconstructed problem expose the factual inconsistency in the original solutions. To rectify the solution, RCoT formulates detected factual inconsistency into fine-grained feedback to guide LLMs in revising solutions. 
  Experimental results demonstrate improvements of RCoT over standard CoT, Self-Consistency and Self-Refine across seven arithmetic datasets. Moreover, we find that manually written fine-grained feedback can dramatically improve LLMs' reasoning abilities (e.g., ChatGPT reaches $94.6\%$ accuracy on GSM8K), encouraging the community to further explore the fine-grained feedback generation methods.
  
\end{abstract}

\section{Introduction}

Large language models (LLMs) \citep{gpt3,zhang2022opt,narang2022pathways,touvron2023llama} have showcased strong reasoning capabilities using chain-of-thought (CoT) \citep{wei2023chainofthought,chowdhery2022palm,NormSage2022}, where LLMs are prompted to generate intermediate steps before the final answer. Despite the impressive performance of CoT prompting across various reasoning tasks \citep{dua2019drop,miao-etal-2020-diverse,cobbe2021,yu2020reclor,bhagavatula2019abductive,talmor-etal-2019-commonsenseqa}, LLMs still struggle to maintain factual consistency in reasoning. Specifically, each reasoning problem usually consists of several conditions and a question, and LLMs exhibit tendencies to \voneh{}, \vtwo{} conditions  and \vthree{} questions \citep{golovneva2022roscoe}. 

\begin{figure}[t]
\centering
\includegraphics[scale=0.43]{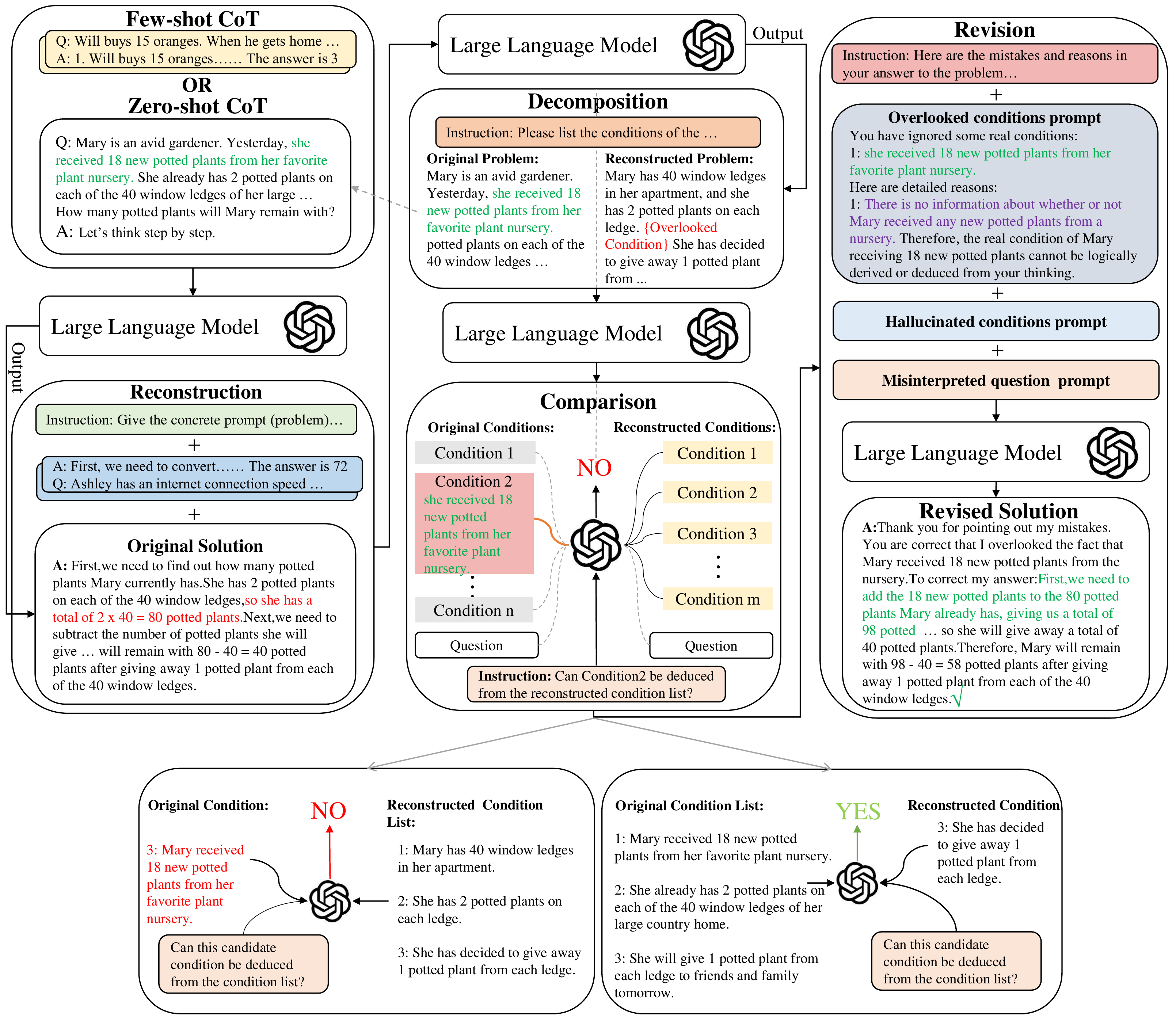}
\caption{The framework of RCoT. (1) \textbf{Reconstruction:} Ask LLMs to reconstruct the problem according to the original solution with instruction and demonstration examples. (2) \textbf{Decomposition:} Decomposing the original problem and reconstructed problem into fine-grained condition lists. (3) \textbf{Comparison:} Compare both lists of sub-conditions and questions to verify whether there are \probone{}s, \probtwo{}s and \probthree{}s.(4) \textbf{Revision:} Gathering all factual inconsistencies into fine-grained feedback to instruct LLMs to revise solutions.}

\label{figure4}
\vspace{-6mm}
\end{figure}

While previous research has proposed various methods to enhance Chain-of-Thought performance~\citep{zhang2022automatic,fu2022complexity,diao2023active,shum2023automatic,zhou2023leasttomost,wang2023selfconsistency,gao2023pal, chen2022program,weng2023large,paul2023refiner,shinn2023reflexion}, there remains a noticeable absence of explicit studies addressing the issue of factual inconsistency. The most relevant work is probably Self-Verification \citep{weng2023large}, which verifies answers by swapping conditions and answers. However, it can only tell whether answers are correct and fail to give fine-grained feedback on factual inconsistency to guide LLMs in revising solutions. Figure~\ref{fig:example_of_factual_inconsistency} shows an instance of factual inconsistent solutions generated by ChatGPT, where the critical condition of "2 days away" is mistakenly \edtwo{}. Despite the incorrect solution, ChatGPT could be rectified if we explicitly point out that it \edtwo{} the "tomorrow, 10/16/1924." condition. Motivated by this observation, we introduce Reverse Chain-of-Thought (RCoT), a novel method for detecting and rectifying condition \probone{} (Figure \ref{hallucinated_example}), \probtwo{} (Figure \ref{fig:example_of_factual_inconsistency})and  question \probthree{} (Figure \ref{misinterpreted_example}) in CoT. More examples could be found in Appendix \ref{app:rcotexample}. The schematics of our proposed method are illustrated in Figure \ref{figure4}. To detect factual inconsistency, RCoT begins with reconstructing new problems from the original solutions. Next, RCoT detects inconsistencies by conducting a fine-grained comparison between the reconstructed problems and the original problems. To rectify inconsistencies, the detected inconsistencies will be formulated as fine-grained feedback to guide the LLMs in revising their solutions. Fine-grained feedback can not only be used to revise solutions but can also offer more interpretability of reasoning errors, allowing humans to inspect solutions easier.

\begin{center}
\begin{tcolorbox}[fonttitle=\bfseries]

\normalsize\textbf{Problem:}
\small{Today's meeting is rescheduled to 11 am \textcolor{red}{tomorrow, 10/16/1924.} What is the date one year ago from today?}\\
\normalsize\textbf{ChatGPT Chain-of-Thought:} \\\small{The current date is 10/16/1924. To find the date one year ago from today, you would subtract one year from the current year, which would be 1923. The correct answer is 10/16/1923.  \textcolor{red}{\ding{55}}}
\end{tcolorbox}
\captionof{figure}{A example of factual inconsistency (condition overlooking).}
\label{fig:example_of_factual_inconsistency}
\end{center}

\begin{center}
\begin{tcolorbox}[title={\textcolor{cyan}{Problem:} The cave is 1218 feet deep and they are already at 849 feet. If they are travelling at speed of 17 \textcolor{darkpastelgreen}{How much farther until they reach the end of the cave?}}]
\normalsize\textbf{ChatGPT Chain-of-Thought:}\\
\small{First, we need to find out how much deeper the cave is from their current position ...... \textcolor{darkpastelgreen}{So they still need to descend 369 feet to reach the end of the cave.} Next, we need to figure out how long it will take them to descend that distance at a speed of 17 feet per unit of time ...... \textcolor{red}{Therefore, it will take them 22 units of time to reach the end of the cave.}  \textcolor{red}{\ding{55}}}
\end{tcolorbox}
\captionof{figure}{An example showing that LLMs may \vthree{} questions. \textcolor{darkpastelgreen}{Green}: The correct solution of the problem. \textcolor{red}{Red}: The superfluous answer given by LLMs. \textbf{Misinterpreting:} LLMs have correctly answered the question in the problem, but it continues answering "how long", resulting in a wrong answer. This is a typical misinterpretation of the problem.}
\label{misinterpreted_example}
\end{center}
\begin{center}
\begin{tcolorbox}[title={\textcolor{cyan}{Problem:} A wooden bridge can carry no more than 5000 pounds. A delivery truck filled with identical boxes, \textcolor{red}{each weighing 15 pounds}, will pass over the bridge. The combined weight of the driver and the empty truck is 3755 pounds. What is the maximum number of boxes which can be loaded onto the truck while not exceeding the bridge's weight limit?}]
\normalsize\textbf{ChatGPT Chain-of-Thought:} \\\small{First, we need to subtract the weight of the driver and the empty truck from the maximum weight the bridge can carry ...... \textcolor{red}{Let's assume each box weighs 16 pounds (15 pounds for the contents and 1 pound for the box).} ...... This means we can load a maximum of 77 boxes onto the truck without exceeding the bridge's weight limit.\textcolor{red}{\ding{55}}}
\end{tcolorbox}
\captionof{figure}{An example showing that LLMs \voneh{} conditions. \textcolor{red}{Red}: Hallucinated conditions. \textbf{Hallucinating:} The problem mentions that each box weighs 15 pounds. However, LLMs assume each
box weighs 16 pounds, which contradicts real conditions.}
\label{hallucinated_example}
\end{center}

We evaluate \ours{} on seven arithmetic reasoning datasets, including GSM8k \citep{cobbe2021training}, AQuA \citep{ling-etal-2017-program}, SVAMP \citep{patel-etal-2021-nlp}, AddSub \citep{hosseini-etal-2014-learning}, ASDiv \citep{miao2021diverse}, Date \citep{srivastava2022imitation} and SingelEq \citep{koncel-kedziorski-etal-2016-mawps}. Experimental results demonstrate the effectiveness of \ours{}, outperforming competitive baselines in both zero-shot and few-shot settings. In-depth analysis and human evaluation suggest that fine-grained feedback on factual inconsistency is crucial for LLMs to revise solutions for arithmetic problems. For example, ChatGPT could achieve $94.6$\% accuracy on GSM8k with manually written fine-grained feedback. Moreover, we conduct comprehensive ablation studies examining the impact of individual modules. Our findings encourage the community to further explore detecting and rectifying factual inconsistency to enhance LLMs' reasoning ability. 

Our contributions are summarized as follows: 
\begin{itemize}
    \item We propose a novel prompting method, \oursfull\ (\ours{}) to effectively detect and rectify the factual inconsistency of LLMs in arithmetic reasoning, focusing on \edtwo{}, \edone{} conditions and \edthree{} questions. RCoT demonstrates improvement over competitive baseline models across seven arithmetic reasoning tasks.
    \item Prompting with fine-grained feedback 
    on factual inconsistency shows encouraging results on improving LLMs' reasoning abilities. Though automatically generated feedback by RCoT shows consistent improvement compared with standard CoT, we find that human-written ground-truth feedback can further improve the LLMs' reasoning ability (e.g., ChatGPT reaches $94.6$\% accuracy on GSM8k). The gap between RCoT's feedback and human-written feedback encourages the community to further explore the automatic generation of fine-grained feedback.
    \item RCoT offers more interpretability to the reasoning errors with fine-grained feedback on factual inconsistency, allowing humans to inspect solutions easier.
\end{itemize}

\section{Related Work}
\label{gen_inst}
\paragraph{Language Model for Reasoning}
Reasoning ability is a critical skill  to solve complex problems, such as arithmetic reasoning \citep{koncel-kedziorski-etal-2016-mawps,roy2016solving,miao-etal-2020-diverse,cobbe2021,dua2019drop}, logical reasoning \citep{yu2020reclor}, commonsense reasoning \citep{bhagavatula2019abductive,talmor-etal-2019-commonsenseqa,zellers-etal-2018-swag,ye2022unreliability}
, and tabular reasoning \citep{zhu2021tat}.
Recently, Large Language Models (e.g., GPT3 \citep{brown2020language}, ChatGPT, PaLM \citep{narang2022pathways} and LLaMA \citep{touvron2023llama}) have demonstrated promising reasoning capability with Chain-of-Thought methods. However, large language models exhibit tendencies to generate intermediate steps that are factually inconsistent, rendering them incapable of solving complex problems requiring multi-step reasoning. In this work, we focus on the detection and rectification of factually inconsistent errors in the intermediate reasoning steps, including question \probthree{}, condition \probone{} and condition \probtwo{}.

\paragraph{Prompt Engineering}
Some prompting methods can elicit useful knowledge in large language models to better solve complex tasks, two representative examples of which are In-context Learning \citep{brown2020language} and Chain-of-Thought \citep{wei2023chainofthought}. In-Context Learning encourages the language models to learn from a few input-output examples as prompts \citep{liu-etal-2022-makes,rubin2022learning,min-etal-2022-metaicl}. Chain-of-Thought prompting improves reasoning performance by prompting LLMs to think of intermediate steps. Inspired by the promising performance of CoT, many methods have explored how to further improve standard CoT. Least-to-most \citep{zhou2023leasttomost} prompting proposes to decompose a complex problem into a series of subproblems and solve them sequentially. Self-Consistency prompting \citep{wang2023selfconsistency} improves performance through majority voting on multiple solutions. Similarly, Complex CoT \citep{fu2022complexity} emphasizes the importance of prompt complexity and selects the most complex examples as prompts. Auto-CoT \citep{zhang2022automatic} is proposed to reduce the workload of manual labeling. Active Prompting \citep{diao2023active} selects the most uncertain questions as demonstration examples to further improve performance. However, these methods fail to address the factual inconsistency problem. Probably the most relevant work are Self-Verification \citep{wang2023selfconsistency}, REFINER \citep{paul2023refiner}, and Reflexion \citep{shinn2023reflexion}. These approaches focus on correcting LLMs outputs. However, Self-Verification can only generate binary feedback and fail to get fine-grained feedback, REFINER needs externally trained models, and Reflexion requires environmental feedback, which cannot be easily obtained in arithmetic reasoning. Compared to these methods,  RCoT entirely relies on the LLM itself to generate fine-grained feedback on factual consistency.

\paragraph{Reverse Engineering.}
\ours{} is inspired by the concept of Reverse Engineering, which has various applications in machine learning research. \citep{fredrikson2014privacy} proposes a reverse method for linear models to evaluate models' privacy safety. \citep{fredrikson2015model} introduces a model inversion method for shallow neural networks, which can reconstruct the face information. \citep{geva2022transformer} unveils the internal prediction construction process of Transformer-based language models by reverse engineering the operations of the feed-forward network (FFN) layers.
Inverting model hyperparameters is another application of reverse engineering techniques. \citep{bhagavatula2019abductive} reverses network parameters by repeatedly requesting the predicted label from the target model. \citep{tramer2016stealing} develops an avatar method to estimate training data and model architectures, while \citep{oh2019towards} trains a set of white-box models to estimate model hyperparameters. \citep{hua2018reverse} estimates both the structure and the weights of a CNN model on a hardware accelerator from information leaks of memory access patterns. 
Different from their goal of opening up the black-box of deep learning models, our work focuses on automatically detecting and rectifying factual inconsistencies that appeared in the solutions generated by LLMs.

\section{\oursfull{} (\ours{})}
\label{headings}

We propose RCoT for detecting and rectifying factual inconsistency (i.e., condition \probone{}s, \probtwo{}s, and question \probthree{}) in CoT to enhance LLMs' reasoning ability. 
Specifically, given a complex reasoning problem $Q$ and original solution $c$ generated by the LLM, we first ask LLMs to detect  factual inconsistency: (i) \textbf{Problem Reconstruction:} Reconstruct the problem $\hat Q$ based on the generated solution $c$. (ii) \textbf{Fine-grained Comparison:} Conduct a fine-grained comparison between the original problem $Q$ and the reconstructed problem $\hat Q$ to detect condition \probone{}s, \probtwo{}s, and question \probthree{}. Then we rectify LLMs using detected factual inconsistency: (iii) \textbf{Fine-grained Feedback and Revision:} The fine-grained comparison reveals the factual inconsistency in original solutions. The detected factual inconsistencies are formulated into fine-grained feedback to guide LLMs in revising their solution accordingly. The overall schematic illustrations of our proposed approach are illustrated in Figure \ref{figure4}, and an example of RCoT is shown in Appendix \ref{app:whole}.

\subsection{Problem Reconstruction}
Intuitively, if the generated step-by-step solution of an arithmetic problem is logically and factually correct and complete, it is more likely for a human to infer what is the original problem. Similarly, we ask the LLM to reconstruct the problem to get $\hat Q$ based on its own solution $c$, in order to verify whether it truly understands the problem. We manually write instructions and in-context examples as the reconstruction prompt.
We find that the factual inconsistencies such as \textit{condition \probone{}s} (e.g., the LLM uses conditions that are not mentioned in the problem $Q$), \textit{condition \probtwo{}s} (e.g., the LLM overlooks some important conditions in the problem $Q$), and \textit{question \probthree{}s} (e.g., the LLM misunderstand the question of $Q$) can be effectively exposed by comparing the reconstructed problem $\hat Q$ with the original problem $Q$ (\S~\ref{subsec:method_comparison}), as shown in Figures \ref{hallucinate2}, \ref{overlook3}, and \ref{misunderstand1} in Appendix \ref{app:rcotexample}, respectively. The prompt template can be found in Figure \ref{prompt_temp}. 

\subsection{Fine-grained Comparison}
\label{subsec:method_comparison}
To detect condition \probone{}s and \probtwo{}s, as well as question \probthree{}s in the solution $c$ from the reconstructed problem $\hat Q$, a naive approach is to ask the LLM to directly compare $Q$ with $\hat Q$. However, such comparisons usually fail to produce high-quality detection results (Figure \ref{figure5}), which is unsurprising because $Q$ and $\hat Q$ contain rich information, and the coarse-grained comparison will inevitably ignore some vital information, causing a sub-optimal result. Therefore, we use fine-grained step-by-step comparisons to improve the detection quality.  All prompt templates are shown in Figure \ref{prompt_temp}. The process is as follows:

\textbf{Problem Decomposition}. $Q$ and $\hat Q$ are unstructured texts, which are hard to be compared in an organized manner. To overcome this issue, we ask the LLM to decompose the problem into a list of conditions $L_Q = [L_Q^1, \cdots, L_Q^m], L_{\hat Q} = [L_{\hat Q}^1, \cdots, L_{\hat Q}^n]$. The structured condition list will then be used in fine-grained comparison. 

\textbf{Condition Comparison} To find the differences between $Q$ and $\hat Q$, we first check whether their condition lists $L_Q$ and $L_{\hat Q}$ are the same. Specifically, the LLM is required to answer whether each $L_Q^i$ can be inferred from $L_{\hat Q}$. If $L_Q^i$ cannot be inferred from $L_{\hat Q}$, then $L_Q^i$ is either (1) \edtwo{} in the solution or (2) \edone{} by the LLM as a different condition. Similarly, we ask the LLM to tell whether $L_{\hat Q}^j$ can be inferred from $L_{Q}$ for every $j$. If $L_{\hat Q}^j$ cannot be inferred from $L_Q$, then $L_{\hat Q}^j$ is \edone{}. Apparently, we need to conduct comparisons for $nm$ times in total. 

\textbf{Question Comparison} The LLM sometimes will also \vthree{} the question (Figure \ref{misinterpreted_example}). Therefore, we also ask LLM to compare the questions being asked in $Q$ and $\hat Q$. If LLMs find the two questions are different, then LLMs  \vthree{} the question in their solutions. This comparison only needs to be done once since there is one question per problem in most cases.

After these comparisons, we detect \edone{} conditions, \edtwo{} conditions, and \edthree{} of questions. We then use them to formulate our fine-grained feedback to guide the LLM in revising its solution.

\subsection{Fine-grained Feedback and Revision}

We assume the original solution is correct if we do not detect any factual inconsistency through fine-grained comparison. On the contrary, we formulate fine-grained feedback to guide the LLM in revising its solution if any factual inconsistency is detected. Specifically, the fine-grained feedback will first state that the solution is incorrect, then list the detected factual inconsistency, and finally ask the LLM to \word{} its solution. Figure \ref{prompt_temp} shows the template we use to formulate the feedback. We take the answer of the \word{}d solution as the final output for evaluation.

\section{Experiment}
\label{headings}
Our extensive experiments aim to show that (1) RCoT benefits arithmetic reasoning by automatically detecting and rectifying condition \probone{} and \probtwo{}, and question \probthree{}; (2) fine-grained feedback of factual consistency is critical for LLMs to self-revise the solution. (3) fine-grained comparison is essential for constructing high-quality fine-grained feedback.

\subsection{Experiment Setting}
\label{experiment setting}
We used closed-source ChatGPT and open-source LLaMA-13B-Chat \citep{touvron2023llama} as the backbone LLMs for solution generation and set the temperature to 0 to improve reproducibility. We evaluate RCoT on seven arithmetic datasets with different difficulties, including GSM8k \citep{cobbe2021training}, AQuA \citep{ling-etal-2017-program}, SVAMP \citep{patel-etal-2021-nlp}, AddSub \citep{hosseini-etal-2014-learning}, ASDiv \citep{miao2021diverse}, Date \citep{srivastava2022imitation} and SingelEq \citep{koncel-kedziorski-etal-2016-mawps}. Due to the high time cost of API calls, we do not use the whole test set but randomly sample test sub-sets. To reduce the randomness caused by test set sampling and make our results more convincing, we sample three test sub-sets that each contains 256 inputs. We report the average accuracy with deviation on the three test sub-sets. For the dataset that has less than 256 test inputs, we still evaluate three times since ChatGPT's outputs may change and report the average accuracy with deviation. A detailed description of each dataset is shown in Appendix \ref{app:dataset}.

We consider both zero-shot and few-shot settings. For the zero-shot setting, we add the prompt "Let’s think step by step" to encourage LLMs to think intermediate steps but without any demonstration examples \citep{kojima2023large}. For the few-shot setting, we use four-shot CoT prompts that consist of problems, solutions, and final answers.

We compare our method with five baselines: (1) \textbf{Chain-of-thought (CoT)} \citep{wei2023chainofthought} (2) \textbf{Active-Prompting} \citep{diao2023active}, a method that selects the most uncertainty problems as demonstration examples. (3) \textbf{Double-Check} asks LLMs to check their answers but does not point out whether the answer is correct. In our experiment, we use the prompt "You should double-check your answers". (4) \textbf{Self-Consistency (SC)} \citep{wang2023selfconsistency} through majority voting on multiple solutions to improve the performance. (5) \textbf{Self-Refine} \citep{madaan2023selfrefine} uses iterative feedback and refinement to revise the answer. We use Tiktoken from Openai to calculate the cost of average tokens.\footnote{https://github.com/openai/tiktoken}

\begin{table}[t]
\renewcommand{\arraystretch}{1.5}
\captionof{table}{Average accuracy and standard deviation on seven arithmetic reasoning datasets. \textbf{Bold} denotes the best result. \textcolor{darkpastelgreen}{Green}: The performance improvement compared with Standard CoT and Active-Prompting in Zero-shot and Few-shot settings, respectively. * denotes the LLM that uses Manual-CoT. \textbf{-} denotes that Active-Prompting \citep{diao2023active} does not support the dataset in their source codes.}
\label{table2}
\vspace{-3mm}
\centering
\resizebox{\linewidth}{!}
{
\begin{tabular}{llccccccc}
\toprule
\multicolumn{1}{l}{\multirow{2}{*}{Model}} & \multicolumn{1}{l}{\multirow{2}{*}{Method}} & \multicolumn{7}{c}{Arithmetic} \\
\cline{3-9}
\multicolumn{1}{c}{}       & \multicolumn{1}{c}{}     & GSM8K                  & AQuA                   & AddSub                  & Date                         &SingleEq                           &ASDiv         &SVAMP             \\ \midrule
UL2-20B$^*$       & Standard              & 4.4                    & 23.6                   & 18.2                    & 14.4                      & 20.2                   & 16.9                                     & 12.5                                                    \\
LaMDA-137B$^*$     & Standard             & 14.3                   & 20.6                   & 51.9                    & 26.8                      & 58.7                   & 46.6                                    & 37.5                  \\                                          
Text-davinci-002$^*$     & Standard       &46.9                    &24.8                    & 81.3                    & 52.1                      &86.6                    &71.3                                     & 68.9 
\\\hline
\multicolumn{9}{c}{\textbf{Zero-shot CoT}}                                                                                                                                                                               \\ \midrule
\multirow{3}{*}{ChatGPT} & Standard                                    & $79.0_{\pm0.95}$            & $51.3_{\pm0.6}$          & $85.2_{\pm1.2}$                            & $66.7_{\pm1.4}$               & $90.3_{\pm0.6}$                   & $84.3_{\pm0.4}$     & $76.7_{\pm4.1}$     \\ \cline{2-9}
 & +Double-Check                                & $79.3_{\pm2.1}$           & $42.7_{\pm0.6}$          & $85.6_{\pm1.2}$                            & $60.5_{\pm6.5}$               & $88.8_{\pm0.8}$                   & $82.8_{\pm1.4}$          & $77.6_{\pm2.0}$                    \\ \cline{2-9}
& +RCoT                                 & $\mathbf{82.0_{\pm0.3}}$  & $\mathbf{55.5_{\pm0.8}}$ & $\mathbf{87.1_{\pm1.1}}$                   & $\mathbf{71.7_{\pm1.3}}$      & $\mathbf{91.4_{\pm0.8}}$          & $\mathbf{86.0_{\pm0.3}}$     & $\mathbf{79.6_{\pm4.1}}$                             \\ \cline{1-9}
                                           &  & \textcolor{darkpastelgreen}{$(+3.1_{\pm0.6})$}                 & \textcolor{darkpastelgreen}{$+(4.1_{\pm0.2})$}      & \textcolor{darkpastelgreen}{$+(1.8_{\pm0.1})$} & \textcolor{darkpastelgreen}{$+(5.0_{\pm0.4})$}  & \textcolor{darkpastelgreen}{$+(1.1_{\pm0.2})$}          & \textcolor{darkpastelgreen}{$+(1.7_{\pm0.3})$}   & \textcolor{darkpastelgreen}{$+(2.8_{\pm0.2})$}  \\ \midrule
\multirow{3}{*}{LLaMA-13B-Chat} & Standard                                    & $36.9_{\pm0.8}$            & $27.2_{\pm0.0}$          & $66.7_{\pm0.5}$                            & $52.4_{\pm1.5}$               & $62.6_{\pm2.6}$                   & $52.2_{\pm3.7}$     & $38.6_{\pm1.1}$     \\ \cline{2-9}
 & +Double-Check                                & $35.6_{\pm1.1}$           & $24.8_{\pm0.0}$          & $62.0_{\pm0.7}$                            & $27.0_{\pm0.9}$               & $62.1_{\pm3.2}$                   & $53.2_{\pm3.6}$          & $41.1_{\pm0.2}$                    \\ \cline{2-9}
& +RCoT                                 & $\mathbf{39.8_{\pm0.8}}$  & $\mathbf{31.9_{\pm0.0}}$ & $\mathbf{67.4_{\pm0.5}}$                   & $\mathbf{55.3_{\pm2.0}}$      & $\mathbf{63.5_{\pm2.1}}$          & $\mathbf{53.0_{\pm3.7}}$     & $\mathbf{41.1_{\pm0.8}}$                             \\ \cline{1-9}
                                           &  & \textcolor{darkpastelgreen}{$(+2.9_{\pm0.4})$}                 & \textcolor{darkpastelgreen}{$+(4.7_{\pm0.0})$}      & \textcolor{darkpastelgreen}{$+(0.7_{\pm0.3})$} & \textcolor{darkpastelgreen}{$+(2.9_{\pm1.0})$}  & \textcolor{darkpastelgreen}{$+(0.9_{\pm0.5})$}          & \textcolor{darkpastelgreen}{$+(0.8_{\pm0.0})$}   & \textcolor{darkpastelgreen}{$+(2.5_{\pm0.4})$}  \\ \midrule

\multicolumn{9}{c}{\textbf{Few-shot CoT}}                                                                                                                 \\ \midrule
\multirow{3}{*}{ChatGPT} & Active-Prompting                          & $81.8_{\pm0.6}$                      & $53.3_{\pm0.6}$                        & $87.2_{\pm1.2}$                           & -            & $91.7_{\pm0.4}$            & $87.9_{\pm0.8}$    & $82.5_{\pm0.6}$  \\  \cline{2-9}
& +Double-Check                               & $77.8_{\pm0.7}$                      & $26.3_{\pm0.5}$                        & $86.0_{\pm1.6}$                           & -            & $91.5_{\pm0.2}$            & $85.7_{\pm2.4}$    & $82.2_{\pm0.8}$     \\   \cline{2-9}
& +RCoT                                & $\mathbf{84.6_{\pm0.6}}$             & $\mathbf{57.1_{\pm0.3}}$               & $\mathbf{88.2_{\pm1.5}}$                  & -            & $\mathbf{93.0_{\pm0.8}}$   & $\mathbf{89.3_{\pm0.5}}$    & $\mathbf{84.9_{\pm1.3}}$ \\ \cline{1-9}
                                            & & \textcolor{darkpastelgreen}{$(+2.7_{\pm0.1})$}                 & \textcolor{darkpastelgreen}{$+(3.7_{\pm0.9})$}      & \textcolor{darkpastelgreen}{$+(1.0_{\pm0.4})$} & -  & \textcolor{darkpastelgreen}{$+(1.2_{\pm0.4})$}              & \textcolor{darkpastelgreen}{$+(1.4_{\pm0.5})$} & \textcolor{darkpastelgreen}{$+(2.3_{\pm1.0})$}\\ \cline{1-9}

\multirow{3}{*}{LLaMA-13B-Chat} & Active-Prompting                          & $37.9_{\pm0.6}$                      & $29.1_{\pm0.0}$                        & $68.4_{\pm0.7}$                           & -            & $67.9_{\pm2.2}$            & $53.3_{\pm0.6}$    & $49.4_{\pm0.4}$  \\  \cline{2-9}
& +Double-Check                               & $36.2_{\pm0.1}$                      & $23.2_{\pm0.0}$                        & $61.9_{\pm2.1}$                           & -            & $64.9_{\pm1.3}$            & $50.3_{\pm3.5}$    & $47.4_{\pm0.8}$     \\   \cline{2-9}
& +RCoT                                & $\mathbf{40.1_{\pm0.4}}$             & $\mathbf{30.7_{\pm0.0}}$               & $\mathbf{68.8_{\pm0.9}}$                  & -            & $\mathbf{68.1_{\pm2.3}}$   & $\mathbf{53.6_{\pm0.4}}$    & $\mathbf{51.2_{\pm0.3}}$ \\ \cline{1-9}
                                            & & \textcolor{darkpastelgreen}{$(+2.1_{\pm0.3})$}                 & \textcolor{darkpastelgreen}{$+(1.6_{\pm0.0})$}      & \textcolor{darkpastelgreen}{$+(0.4_{\pm0.3})$} & -  & \textcolor{darkpastelgreen}{$+(0.2_{\pm0.1})$}              & \textcolor{darkpastelgreen}{$+(0.3_{\pm0.2})$} & \textcolor{darkpastelgreen}{$+(1.8_{\pm0.2})$}\\ \bottomrule
\end{tabular}
}
\end{table}

\subsection{RCoT benefits arithmetic reasoning}
Table \ref{table2} shows the results of RCoT on seven arithmetic datasets. Our method consistently outperforms the standard CoT and the double-check methods in the zero-shot setting. Moreover, LLMs benefit more from our method on more challenging tasks that require complex reasoning. For example, the AQuA dataset contains diverse problems, and
the Date dataset requires multi-hop reasoning and common sense date knowledge. Both ChatGPT and LLaMA achieve  lower accuracy scores on AQuA and Date (51.3\% and 66.7\% for ChatGPT and 27.2\% and 52.4\% for LLaMA) among all seven tasks. 
Meanwhile, we observe that our method helps LLMs improve by apparent margins on AQuA and Date (4.1\%, 5.0\% and 4.7\%, 2.9\% for ChatGPT and LLaMA), the highest gains in all seven tasks. Our method also remains effective for easier tasks. For example, RCoT enhances the performance of the SVAMP dataset, which contains problems that usually only need one-step calculation, by 2.8\% and 2.5\%. Moreover, we also observe greater improvements from our method on ChatGPT than LLaMA, potentially due to the stronger abilities of ChatGPT to detect and correct errors.

We can observe similar results in the few-shot setting to the zero-shot setting. Although selecting the most uncertain problems for LLMs as demonstration examples is helpful for reasoning \citep{diao2023active}, RCoT still improves the accuracy. It is worth noting that the performance of Double-Check method in the few-shot CoT setting decreases immensely. On the AQuA and GSM8K datasets, its performance drops by 27.0\% and 4.0\%, suggesting that few-shot examples may increase the risk of revising correct solutions to the incorrect ones.  LLaMA exhibits a lower degree of susceptibility compared to ChatGPT.

We also compared RCoT with other stronger baselines (i.e., Self-Consistency, SC for short, and Self-Refine). Specifically, We conducted 30 trials per problem for SC and 3 trials per problem for RCoT in the zero-shot setting (set temperature to 0.7 \citet{wang2023selfconsistency}), which uses similar costs. Due to the extremely high cost, we do not experiment with the few-shot setting and leave it as our future work. We set max attempt to 5 for Self-Refine.
Table \ref{table4} has shown the results. RCoT could achieve comparable performance to SC at nearly one-third of the cost (e.g., AddSub, SingleEq, SVAMP) and even outperforms SC on the GSM8K dataset. However, the performance significantly drops on AQuA and Date datasets. That is because there are multiple-choice tasks, making it exceedingly simple for the model to approximate the answer by employing multiple guesses with incorrect logical steps. Combining RCoT with SC, our method can further improve the performance and surpass all baselines, reaching a high accuracy of 84.5\%  across seven arithmetic datasets. Our experiments demonstrate the same conclusion as \citet{madaan2023selfrefine} that Self-Refine is not good at arithmetic reasoning. It's worth noting that Self-Refine achieves the highest accuracy on SingleEq and AddSub datasets. Nevertheless, the improvement does not come from refinement but the usage of code in the Self-Refine implementation, reducing a large number of calculation errors. The real improvements brought by refinement are actually 0.8 and 0.4 in the AddSub and SingleEq datasets, respectively. Another phenomenon is that self-refine does not bring more token cost even if we give it more refinement budget. This is because self-refine tends to state that the solution is correct after the second refinement.

\begin{table}[!ht]
\centering
\begin{minipage}[c]{0.48\textwidth}
\caption{The performance of RCoT using fine-grained feedback and coarse-grained feedback. \textbf{w/o reasons :} remove explanations of specific mistakes from the original fine-grained feedback. The prompt becomes "Your answer is wrong. You should double-check your answer.". \textbf{w/o judgment+reasons:} further remove the high-level judgment. The prompt becomes "You should double-check your answer." \textcolor{red}{Red}: The performance drops compared with RCoT method.}
\label{table3}
\vspace{1mm}
\centering
\renewcommand{\arraystretch}{1.67}
\setlength{\tabcolsep}{3pt}
\scalebox{0.75}{
  \begin{tabular}{lccc}
    \toprule
    Method                      & GSM8K         & AQUA      & SVAMP                     \\
    \midrule
    Standard CoT                   & 79.0          & 51.3      & 76.7                      \\
    \midrule
    RCoT(ours)           & \textbf{82.0}          & \textbf{55.5}      & \textbf{79.6}                        \\
     \ \ - w/o reasons       & 80.0 \textcolor{red}{(-2.0)}          & 52.3 \textcolor{red}{(-3.2)}     & 78.9 \textcolor{red}{(-0.7)}                      \\
     \ \ - w/o judgment+reasons                & 79.3 \textcolor{red}{(-2.7)}         & 42.7 \textcolor{red}{(-12.8)}     & 77.6 \textcolor{red}{(-2.0)}                      \\
    \bottomrule
  \end{tabular}}
\end{minipage}
\hspace{3mm}
\begin{minipage}[c]{0.48\textwidth}
\caption{The performance without question comparison and condition comparison, as well as the performance with coarse-grained comparison. \textbf{coarse-grained:} We directly ask LLMs to compare the original problem with the reconstructed problem. Results show that (1) fine-grained comparison is important to get fine-grained feedback, and (2) both question comparison and condition comparison are important in the fine-grained comparison. \textcolor{red}{Red}: The performance drops compared with the RCoT method. }
\label{table4}
\vspace{1mm}
\centering
\renewcommand{\arraystretch}{1.25}
\setlength{\tabcolsep}{3pt}
\resizebox{\textwidth}{!}{
  \begin{tabular}{lccc}
    \toprule
    Method                      & GSM8K                                 & AQUA                              & SVAMP                     \\
    \midrule
    Standard CoT                   & 79.0                                  & 51.3                              & 76.7                      \\
    \midrule
    RCoT           & \textbf{82.0}          & \textbf{55.5}      & \textbf{79.6}                        \\
     \ \ - w/o question comparison    & 80.9 \textcolor{red}{(-1.1)}        & 54.6 \textcolor{red}{(-0.9)}    & 79.2 \textcolor{red}{(-0.4)}                                             \\
     \ \ - w/o condition comparison       & 80.1 \textcolor{red}{(-1.9)}          & 53.5 \textcolor{red}{(-2.0)}      & 78.1 \textcolor{red}{(-1.5)}                       \\
    RCoT (Corase-grained)              & 74.2 \textcolor{red}{(-7.8)}                                    & 49.6 \textcolor{red}{(-5.9)}                               & 76.1 \textcolor{red}{(-3.5)}                       \\
    \bottomrule
  \end{tabular}}
\end{minipage}
\end{table}

\subsection{Fine-grained feedback is critical for solution revision}

The success of our method comes from fine-grained feedback that points out detailed factual inconsistency (condition \probone{} and \probtwo{}, and question \probthree{}). In this section, we show that coarse-grained feedback will lead to worse performance to prove the necessity of fine-grained feedback. We replace our fine-grained feedback with two kinds of coarse-grained feedback: (1) w/o reasons: we do not tell LLMs the detected factual inconsistency by RCoT and only give a high-level judgment. Therefore, if RCoT detects no factual inconsistency, we take the original solution as the final output for evaluation. Otherwise, we use the prompt "Your answer is wrong. You should double-check your answer" to guide LLMs in revising solutions. (2) w/o judgment+reasons (i.e., Double-Check): We further remove the high-level judgment from the prompts. Therefore, we always use "You should double-check your answer" to guide LLMs in revising solutions regardless of the detection results of RCoT.
\begin{table}[t]
\renewcommand{\arraystretch}{1.3}
\caption{Average accuracy on seven arithmetic reasoning datasets among Self-Consistency \citep{wang2023selfconsistency}, RCoT and Self-Refine \citep{madaan2023selfrefine}. \textbf{Bold} denotes the best result.}
\label{table4}
\vspace{-3mm}
\centering
\resizebox{\linewidth}{!}
{
\begin{tabular}{lcccccccccc}
\toprule
\multicolumn{2}{l}{Method}                      & GSM8K & AQuA  & AddSub & Date  & SingleEq & ASDiv & SVAMP & Avg Acc & Avg Tokens \\ \hline
\multicolumn{2}{l}{SC (30 trials per problem)}  & 81.6 & 70.8 & 88.6  & \textbf{80.0} & 92.9    & 90.2 & 80.4 &83.5 & 5615.0\\ \cline{3-11}
\multicolumn{2}{l}{RCoT (1 trial per problem)}  & 82.0 & 56.3 & 87.2  & 71.9 & 92.4   & 86.3 & 79.7 &79.4 & 1831.0\\ \cline{3-11}
\multicolumn{2}{l}{RCoT (3 trials per problem)} & \textbf{83.2} & \textbf{72.8} & 89.8  & 78.9 & 93.8    & \textbf{91.8} & \textbf{81.2} & \textbf{84.5} & 5453.3\\
\cline{3-11}
\multirow{6}{*}{Self-Refine}     & attempt 0    & 79.1 & 45.2 & 90.6  & 51.3 & 97.6    & 83.5 & 75.2 & 74.7 & 190.2\\ \cline{2-11}
                                 & attempt 1    & 80.7 & 49.2 & \textbf{91.4}  & 52.7 & \textbf{98.0}    & 84.3 & 76.8 & 76.1 & 3108.4\\ \cline{2-11} 
                                 & attempt 2    & 80.7 & 49.2 & 91.4  & 52.7 & 98.0    & 84.3 & 76.8 & 76.1 &3324.9\\ \cline{2-11}
                                 & attempt 3    & 80.7 & 49.2 & 91.4  & 52.7 & 98.0    & 84.3 & 76.8 & 76.1 & 3359.6\\ \cline{2-11} 
                                 & attempt 4    & 80.7 & 49.2 & 91.4  & 52.7 & 98.0    & 84.3 & 76.8 & 76.1 &  3367.7\\ \cline{2-11}

                                & attempt 5  & 80.7 & 49.2 & 91.4  & 52.7 & 98.0    & 84.3 & 76.8 & 76.1 & 3367.7\\ \bottomrule
\end{tabular}
}
\vspace{-3mm}
\end{table}
Table \ref{table3} shows the results on SVAMP(easy), GSM8K(medium), and AQuA (hard) datasets. We can see consistent performance drops when we remove detected factual inconsistency and only keep a high-level judgment, showing the effectiveness of fine-grained feedback. Moreover, we can observe that further removing judgment will make the performance even worse than standard CoT. This is not surprising because LLMs may mistakenly revise the correct solution to the incorrect one. Appendix \ref{app:rcotdc} shows an example of RCoT and Double-Check, where we can see that RCoT could help the LLM correct the solution, but Double-Check couldn't.

To further show the power of fine-grained feedback, we conduct a human evaluation. Specifically, we go through the generated solution and write fine-grained feedback on factual inconsistency by ourselves. Amazingly, the LLM could reach $94.6\%$ accuracy on the GSM8K dataset and could only reach $86.3\%$ if we remove explanations of factual inconsistencies from our feedback (i.e., the same setting as "w/o reasons" in Table \ref{table3}). Appendix \ref{app:humanprompt} shows examples of manually written and RCoT-generated feedback. This result shows the same observation of Table \ref{table3} and reveals the strong power of fine-grained feedback. Since RCoT still has a gap ($12.6\%$ accuracy gap) compared with humans, we encourage the community to further explore fine-grained feedback generation.

\subsection{Fine-grained comparison leads to fine-grained feedback}
To get fine-grained feedback, conditions and questions are compared in a fine-grained manner in RCoT. A simpler way is to ask LLMs to generate fine-grained feedback directly by comparing the original problems and reconstructed problems. Appendix \ref{sec:quality} has illustrated that LLMs can reconstruct questions well when the reasoning maintains factual consistency but not vice versa.
Table \ref{table4} shows that coarse-grained comparison will cause a significant accuracy drop (even worse than standard CoT), which suggests that it fails to generate high-quality feedback (Figure \ref{figure5}). Therefore, problem decomposition and fine-grained comparison are essential. We also show that both condition comparison and question comparison are important. Removing either of them will lead to worse performance. This is because LLMs may \voneh{}/\vtwo{} conditions (Figure \ref{hallucinated_example}, \ref{fig:example_of_factual_inconsistency}) as well as \vthree{} questions (Figure \ref{misinterpreted_example}).

\begin{center}
\begin{tcolorbox}[title = \textcolor{cyan}{Original Problem:} Rebecca wants to \textcolor{red}{split a collection of eggs into groups of 3}. Rebecca has 99 bananas 9 eggs and 27 marbles. How many groups will be created?\\
\textcolor{cyan}{Reconstructed Problem:} Rebecca has 9 eggs\, 99 bananas\, and 27 marbles. \textcolor{red}{If we can group items into sets of 3}\, how many groups can we make in total?]

\normalsize\textbf{Coarse-Grained Comparison:}\\
\small{\textbf{Yes,} the reconstructed problem does not add new conditions or lack any conditions that were present in original problem. Original problem and reconstructed problem essentially ask the same question, with the only difference being the way the information is presented. Both problems ask how many groups of 3 items can be made from a collection of 9 eggs, 99 bananas, and 27 marbles. Therefore, reconstructed problem does not introduce any new conditions or leave out any conditions from the original problem.}  \textcolor{red}{\ding{55}}
\tcblower
\normalsize\textbf{Fine-Grained Comparison:}\\
\small{\textbf{No,} the candidate condition "Group items into sets of 3" cannot be deduced from the given condition list as it is not applicable to all items in the condition list. The condition only applies to the eggs that Rebecca wants to split into groups of 3. It does not apply to the bananas or the marbles.\\
Additionally, the candidate condition implies that all items must be grouped into sets of 3, but there is no indication in the context that Rebecca wants to group all items into sets of 3. She only wants to split the eggs into groups of 3.}  \textcolor{darkpastelgreen}{\ding{52}}
\end{tcolorbox}
\vspace{-3mm}
\captionof{figure}{This is an example showing the failure of the coarse-grained comparison. \textcolor{red}{Red}: The inconsistent condition between the original problem and the reconstructed problem.}
\label{figure5}
\end{center}

\subsection{Additional Quantitative analysis of RCoT}
\begin{wraptable}{r}{0.5\textwidth}
\vspace{-6mm}
\renewcommand{\arraystretch}{0.8}
{
\captionof{table}{\textbf{Found/Not Found:} RCoT can or cannot find the reasons for errors.\textbf{Other errors:} such as computation error, logical error and so on.}
\label{table20}
\resizebox{0.5\textwidth}{!}{
\begin{tabular}{lccc}
\toprule
Type              & Found & Not Found & total \\ \midrule
Overlooking       & 5          & 1                & 6     \\ \midrule
Hallucinating     & 16         & 15               & 31    \\ \midrule
Misinterpreting & 5          & 3                & 8     \\ \midrule
Other errors      & 0          & 55               & 55    \\ \bottomrule
\end{tabular}
}
}
\vspace{-8mm}
\end{wraptable}
To explore the effectiveness of RCoT, we construct further quantitative analysis on 100 problems the ChatGPT answered incorrectly. We manually divided these problems into four categories: condition hallucination, condition overlooking, question misinterpretation and other errors. The statistical results are shown in Table \ref{table20}. We find that RCoT is better at detecting overlooking and misinterpretation errors than  hallucination errors.

\section{Conclusion}
\vspace{-3mm}
In this paper, we propose RCoT, a method that enables LLMs to detect and rectify factual inconsistency automatically to improve LLMs' reasoning abilities. RCoT detects factual inconsistency through fine-grained comparison between the reconstructed problems and original questions, and then asks LLMs to rectify inconsistencies through fine-grained feedback. Experimental results on seven arithmetic reasoning datasets demonstrate the effectiveness of RCoT. Our experiments also show encouraging results of LLMs' reasoning abilities with the help of manually written fine-grained feedback, encouraging the community to further explore fine-grained feedback generation. RCoT could, in principle, be applied to other tasks requiring CoT solutions. We discuss the limitations and future work in Appendix \ref{app:limit}.

\section{Reproducibility Statement}
The supplementary material includes the code for all experiments and their corresponding running scripts. The dataset (GSM8K, AQuA, AddSub, SingleEq, Date, ASDiv and SVAMP) can be easily accessible on the HuggingFace website or from their official repositories. We explain all the experimental details (temperature, dataset size and so on ) in Section \ref{experiment setting}.

\bibliography{references.bib}
\bibliographystyle{iclr2024_conference}

\clearpage
\appendix
\section{The quality of reconstructed problem}
\label{sec:quality}
We measured the Rouge1, Rouge2, RougeL, RougeSum, and sentence embedding similarity (\texttt{using sentence-transformers/all-mpnet-base-v2}) between original problems and reconstructed problems. We can observe from table \ref{rouge_sim} that higher CoT accuracies correspond to higher similarities between original problems and reconstructed problems. This is expected since correctly solved problems' reconstruction should ideally be itself, and incorrectly solved problems' reconstruction should differ from itself.

\begin{table}[ht]
\renewcommand{\arraystretch}{1.3}
{
\captionof{table}{The metrics between original problem and reconstructed problem}
\begin{tabular}{lcccccc}
\toprule
\textbf{Dateset} & \textbf{Rouge1} & \textbf{Rouge2} & \textbf{RougeL} & \textbf{RougeSum} & \textbf{Similarity} & \textbf{Standard CoT Acc} \\ \hline
GSM8K            & 71.4622         & 49.3915         & 58.8991         & 58.8974           & 93.57              & \textbf{79.0}            \\ \hline
AQuA             & 54.2383         & 33.5828         & 43.5771         & 43.7721           & 84.44              & \textbf{51.3}            \\ \hline
AddSub           & 78.0939         & 57.5594         & 66.3973         & 66.4099           & 94.05              & \textbf{85.2}            \\ \hline
Date             & 46.9414         & 28.7005         & 39.2268         & 39.3934           & 79.12              & \textbf{66.7}            \\ \hline
SingleEq         & 72.8212         & 53.5232         & 64.8562         & 64.7956           & 94.01               & \textbf{90.3}            \\ \hline
ASDiv            & 68.1849         & 46.5488         & 59.7907         & 59.7911           & 92.16              & \textbf{84.3}            \\ \hline
SVAMP            & 75.0074         & 55.3162         & 65.1539         & 65.1973           & 93.71              & \textbf{76.7}            \\ \bottomrule
\end{tabular}
}
\label{rouge_sim}
\end{table}

\section{Examples}

\subsection{Examples for different kinds of factual inconsistencies found by RCoT}
\label{app:rcotexample}
\begin{tcolorbox}[fonttitle=\bfseries]
\normalsize\textbf{Original Solution:}\\
\small{First, we need to find out how many potted plants Mary currently has. She has 2 potted plants on each of the 40 window ledges, \textcolor{red}{so she has a total of 2 x 40 = 80 potted plants.}\\
Next, we need to subtract the number of potted plants she will give away. She has decided to give away 1 potted plant from each of the 40 window ledges, so she will give away a total of 1 x 40 = 40 potted plants.\\
Therefore, Mary will remain with 80 - 40 = 40 potted plants after giving away 1 potted plant from each of the 40 window ledges.\textcolor{red}{\ding{55}}}\\
\\
\normalsize\textbf{Original Problem:}
\small{Mary is an avid gardener. \textcolor{red}{Yesterday, she received 18 new potted plants from her favorite plant nursery.} She already has 2 potted plants on each of the 40 window ledges of her large country home. Feeling generous, she has decided that she will give 1 potted plant from each ledge to friends and family tomorrow. How many potted plants will Mary remain with?}\\\\
\normalsize\textbf{Reconstructed Problem:}
\small{Mary has 40 window ledges in her apartment, and she has 2 potted plants on each ledge. She has decided to give away 1 potted plant from each ledge. How many potted plants will Mary have left after giving away the plants?}

\end{tcolorbox}
\captionof{figure}{A example for \probtwo{}}
\label{overlook1}

\begin{tcolorbox}[fonttitle=\bfseries]
\normalsize\textbf{Original Solution:}\\
\small{First, we need to find the date that is one month before the deadline of June 1, 2021. To do this, we can subtract one month from the deadline date. \\
June 1, 2021 - 1 month = May 1, 2021 \\
So, the date that is one month before the deadline is May 1, 2021. \\
Next, we need to write this date in the format MM/DD/YYYY. \\
May 1, 2021 = 05/01/2021 \\
Therefore, the answer is (C) 05/01/2021.\textcolor{red}{\ding{55}}}\\
\\
\normalsize\textbf{Original Problem:}
\small{\textcolor{red}{The deadline is Jun 1, 2021, which is 2 days away from now.} What is the date a month ago in MM/DD/YYYY? Answer Choices: (A) 04/30/2021 (B) 02/19/2021 (C) 05/01/2021 (D) 05/09/2021 (E) 07/06/2021 (F) 08/03/2021}\\\\
\normalsize\textbf{Reconstructed Problem:}
\small{What is the date that is \textcolor{red}{one month before the deadline of June 1, 2021}, written in the format MM/DD/YYYY? Answer Choices: (A) 04/30/2021 (B) 02/19/2021 (C) 05/01/2021 (D) 05/09/2021 (E) 07/06/2021 (F) 08/03/2021}

\end{tcolorbox}
\captionof{figure}{A example for \probtwo{}}
\label{overlook2}

\begin{tcolorbox}[fonttitle=\bfseries]
\normalsize\textbf{Original Solution:}\\
\small{Let's first calculate how much money Tobias saved up in three months:He gets a \$5 allowance per month, so in three months he has saved up:\\
\$5/month * 3 months = \$15\\
n addition to his allowance, Tobias also earns money by mowing lawns andshoveling driveways.If he mows 4 lawns, he earns:\\
4lawns*\$15/lawn = \$50\\
Let's say he shovels x driveways. Then, he earns:\\
x driveways*\$7/driveway = \$7x\\
In total, he has saved up:\\
\$15 (allowance) + \$60 (lawn mowing) + \$7x (shoveling) = \$75 + \$7x\\
He spent \$95 on shoes, so he has:\\
\$75 + \$7x- \$95 = \$0\\
Simplifying this equation:\\
\$7x - \$20 = \$0\\
\$7x = \$20\\
x= 2.85\\
Therefore, Tobias shoveled about 2.85 driveways. However since he can't shovel afraction of a driveway, we can round this up to 3 driveways.\textcolor{red}{\ding{55}}}\\
\\
\normalsize\textbf{Original Problem:}
\small{Tobias is buying a new pair of shoes that costs \$95. He has been saving up his money each month for the past three months. He gets a \$5 allowance a month. He also mows lawns and shovels driveways. He charges \$15 to mow a lawn and \$7 to shovel. \textcolor{red}{After buying the shoes, he has \$15 in change.} If he mows 4 lawns, how many driveways did he shovel?}\\\\
\normalsize\textbf{Reconstructed Problem:}
\small{How many driveways did Tobias have to shovel to save enough money to buy \$95 shoes, if he gets a \$5 allowance per month for three months, mows 4 lawns for \$15 each, and earns \$7 for every driveway he shovels?}
\end{tcolorbox}
\captionof{figure}{A example for \probtwo{}.}
\label{overlook3}

\begin{tcolorbox}[fonttitle=\bfseries]
\normalsize\textbf{Original Solution:}\\
\small{First, we need to find out how far John traveled during the first 3 hours at a speed of 60 mph. We can use the formula distance = speed x time, so:\\
distance = 60 mph x 3 hours = 180 miles\\
Next, we need to find out how far John traveled during the next 2 hours in standstill traffic. Since he didn't move, the distance is 0 miles.\\
Then, we need to find out how far John traveled during the next half-hour at a speed of 30 mph. Using the same formula, we get:\\
distance = 30 mph x 0.5 hours = 15 miles\\
Finally, we need to find out how far John traveled during the last 1.5 hours at a speed of 80 mph. Again, using the same formula, we get:\\
distance = 80 mph x 1.5 hours = 120 miles\\
\textcolor{red}{To find out how far John is from home at the end of those 4 hours, we need to add up all the distances he traveled:\\
180 miles + 0 miles + 15 miles + 120 miles = 315 miles}\\
Therefore, John is 315 miles away from home at the end of those 4 hours.\textcolor{red}{\ding{55}}}\\
\\
\normalsize\textbf{Original Problem:}
\small{John drives for 3 hours at a speed of 60 mph and \textcolor{red}{then turns around because he realizes he forgot something very important at home.}  He tries to get home in 4 hours but spends the first 2 hours in standstill traffic.  He spends the next half-hour driving at a speed of 30mph, before being able to drive the remaining time of the 4 hours going at 80 mph.  How far is he from home at the end of those 4 hours?}\\\\
\normalsize\textbf{Reconstructed Problem:}
\small{John drove for 4 hours. During the first 3 hours, he drove at a speed of 60 mph. During the next 2 hours, he was stuck in standstill traffic. During the next half-hour, he drove at a speed of 30 mph. During the last 1.5 hours, he drove at a speed of 80 mph. How far is John from home at the end of those 4 hours?}

\end{tcolorbox}
\captionof{figure}{A example for \probtwo{}}
\label{overlook4}

\begin{tcolorbox}[fonttitle=\bfseries]
\normalsize\textbf{Original Solution:}\\
\small{First, we need to find out how many books and movies are left unread/watched. \\
Books left unread: 11 - 7 = 4 \\
Movies left unwatched: 17 - 21 = -4 \\
Uh oh, we have a problem. We can't have negative movies left unwatched, so we need to adjust our calculations. \\
\textcolor{red}{Let's assume that the person has actually watched 17 movies, but some of them were repeats. So, we can say that they have watched 17 unique movies and some of them multiple times.} \\
Now, let's recalculate: \\
Books left unread: 11 - 7 = 4 \\
Movies left unwatched: 17 - 17 = 0 \\
So, the person has read 7 books and watched 17 unique movies. To find out how many more movies than books they have watched, we can subtract: \\
17 - 7 = 10 \\
Therefore, the person has watched 10 more movies than books they have read.\textcolor{red}{\ding{55}}}\\
\\
\normalsize\textbf{Original Problem:}
\small{There are 11 different books and 17 different movies in the ' crazy silly school ' series. If you read 7 of the books and \textcolor{red}{watched 21 of the movies} How many more movies than books have you read?}\\\\
\normalsize\textbf{Reconstructed Problem:}
\small{A person has 11 books and 21 movies. They have read/watched 7 books \textcolor{red}{and 17 movies}. How many more movies than books have they watched?}

\end{tcolorbox}
\captionof{figure}{A example for \probone{}.}
\label{hallucinate1}

\begin{tcolorbox}[fonttitle=\bfseries]
\normalsize\textbf{Original Solution:}\\
\small{First, we know that the year is 1969, so we don't need to worry about that.\\ 
Next, we know that it is currently 4/19, \textcolor{red}{but we need to figure out what day it will be tomorrow.} \\
To do this, we add 1 to the day, which gives us 4/20. \\
Therefore, the answer is (B) 04/20/1969.\textcolor{red}{\ding{55}}}\\
\\
\normalsize\textbf{Original Problem:}
\small{It is 4/19/1969 today. What is the date today in MM/DD/YYYY? Answer Choices: (A) 04/18/1969 (B) 04/20/1969 (C) 04/12/1969 (D) 05/17/1969 (E) 04/19/1969 (F) 07/19/1969}\\\\
\normalsize\textbf{Reconstructed Problem:}
\small{What is \textcolor{red}{the date tomorrow if today is April 19, 1969}? Answer Choices: (A) 04/18/1969 (B) 04/20/1969 (C) 04/12/1969 (D) 05/17/1969 (E) 04/19/1969 (F) 07/19/1969}

\end{tcolorbox}
\captionof{figure}{A example for \probone{}.}
\label{hallucinate2}

\begin{tcolorbox}[fonttitle=\bfseries]
\normalsize\textbf{Original Solution:}\\
\small{First, we need to calculate the total number of petals on the flowers that Rose
picked:\\
The 3 flowers with 5 petals each have a total of 3 x 5 = 15 petals.\\
The 4 flowers with 6 petals each have a total of 4 x 6 = 24 petals.\\
The 5 flowers with 4 petals each have a total of 5 x 4 = 20 petals.\\
The 6 flowers with 7 petals each have a total of 6 x 7 = 42 petals.\\
Therefore, the total number of petals on all the flowers that Rose picked is 15 + 24+ 20 + 42 = 101 petals.\\
\textcolor{red}{Next, we need to subtract one petal from each type of flower that she dropped. Sowe need to subtract:\\
1 petal from the 3 flowers with 5 petals each, leaving 3 x 4 = 12 petals.\\
1 petal from the 4 flowers with 6 petals each, leaving 4 x 5 = 20 petals.\\
1 petal from the 5 flowers with 4 petals each, leaving 5 x 3 = 15 petals.\\
1 petal from the 6 flowers with 7 petals each, leaving 6 x 6 = 36 petals.}\\
Therefore, the total number of petals on the flowers that Rose puts in the vase is 12+ 20 + 15 + 36 = 83 petals.\textcolor{red}{\ding{55}}}\\
\\
\normalsize\textbf{Original Problem:}
\small{Rose is out picking flowers for a vase she wants to fill. She starts off by picking 3 flowers with 5 petals each. She then picks 4 flowers with 6 petals each. She then adds another 5 flowers with 4 petals each. Lastly she picks 6 flowers with 7 petals each. As she's carrying these flowers over to fill the vase, \textcolor{red}{she drops 1 of each and the wind blows them away.} She puts the remnaining flowers in the vase.How many petals in total are on the flowers in the vase?}\\\\
\normalsize\textbf{Reconstructed Problem:}
\small{Rose went flower-picking and collected different types of flowers with varying numbers of petals. She picks 3 flowers with 5 petals each, 4 flowers with 6 petals each, 5 flowers with 4 petals each and 6 flowers with 7 petals each. \textcolor{red}{She later dropped one petal from each type of flower and put the remaining flowers in a vase.} What is the total number of petals on the flowers that she put in the vase?}

\end{tcolorbox}
\captionof{figure}{A example for \probone{}}
\label{hallucinate3}

\begin{tcolorbox}[fonttitle=\bfseries]
\normalsize\textbf{Original Solution:}\\
\small{First, we need to find out how many bottle caps and wrappers Danny had before he found the ones at the park. To do this, \textcolor{red}{we can add the number he found at the park to the number he has now: \\
71 + 28 = 99 bottle caps \\
24 + 16 = 40 wrappers} \\
Next, we need to find the difference between the number of bottle caps and wrappers: \\
99 - 40 = 59 \\
Therefore, Danny has 59 more bottle caps than wrappers in his collection now.\textcolor{red}{\ding{55}}}\\
\\
\normalsize\textbf{Original Problem:}
\small{Danny collects bottle caps and wrappers. \textcolor{red}{He found 71 bottle caps and 24 wrappers at the park. Now he has 28 bottle caps and 16 wrappers in his collection.} How many more bottle caps than wrappers does danny have now?}\\\\
\normalsize\textbf{Reconstructed Problem:}
\small{\textcolor{red}{Danny found 28 bottle caps and 16 candy bar wrappers at the park. He now has 71 bottle caps and 24 candy bar wrappers in his collection.} How many more bottle caps than candy bar wrappers does Danny have in his collection now?}

\end{tcolorbox}
\captionof{figure}{A example for \probone{}}
\label{hallucinate4}

\begin{tcolorbox}[fonttitle=\bfseries]
\normalsize\textbf{Original Solution:}\\
\small{\textcolor{red}{One week ago from today would be August 25th, 2021.} Therefore, the answer is not (A) or (C) since they both represent the current date. It is also not (B), (D), or (F) since they are not one week ago from today. The correct answer is (E) 09/07/2021.\textcolor{red}{\ding{55}}}\\
\\
\normalsize\textbf{Original Problem:}
\small{\textcolor{red}{It was Sept.1st, 2021 a week ago.} What is the date one week ago from today in MM/DD/YYYY? Answer Choices: (A) 09/01/2021 (B) 12/06/2021 (C) 09/02/2021 (D) 09/01/2020 (E) 09/07/2021 (F) 09/01/2098}\\\\
\normalsize\textbf{Reconstructed Problem:}
\small{What is the date that was one week ago \textcolor{red}{from today's date, which is September 1st, 2021?} Choose the correct answer from the following options: (A) 09/01/2021 (B) 12/06/2021 (C) 09/02/2021 (D) 09/01/2020 (E) 09/07/2021 (F) 09/01/2098}

\end{tcolorbox}
\captionof{figure}{A example for \probone{}}
\label{hallucinate5}

\begin{tcolorbox}[fonttitle=\bfseries]
\normalsize\textbf{Original Solution:}\\
\small{First, we need to figure out what day was yesterday. \textcolor{red}{To do that, we subtract one day from 11/23/1933, which gives us 11/22/1933.\\
Next, we need to figure out what day is today. To do that, we subtract one more day from 11/22/1933, which gives us 11/21/1933.}\\
Therefore, the answer is (E) 08/29/1933.\textcolor{red}{\ding{55}}}\\
\\
\normalsize\textbf{Original Problem:}
\small{\textcolor{red}{The day before yesterday was 11/23/1933.} What is the date today in MM/DD/YYYY? Answer Choices: (A) 01/04/1934 (B) 11/04/1933 (C) 11/25/1967 (D) 12/09/1933 (E) 08/29/1933 (F) 11/25/1933}\\\\
\normalsize\textbf{Reconstructed Problem:}
\small{\textcolor{red}{If today is 11/23/1933}, what is the date today? Choose the correct answer from the following options: (A) 01/04/1934 (B) 11/04/1933 (C) 11/25/1967 (D) 12/09/1933 (E) 08/29/1933 (F) 11/25/1933}

\end{tcolorbox}
\captionof{figure}{A example for hallucinating}
\label{hallucinate6}

\begin{tcolorbox}[fonttitle=\bfseries]
\normalsize\textbf{Original Solution:}\\
\small{Martha has been collecting shells since she turned 5 years old, which means she has been collecting shells for 5 years. In one year, Martha collects 12 shells (one per month). So in 5 years, Martha collects 5 x 12 = 60 shells. \textcolor{red}{By her 10th birthday, Martha will have collected 60 + 12 = 72 shells.}\textcolor{red}{\ding{55}}}\\
\\
\normalsize\textbf{Original Problem:}
\small{Martha has been collecting shells since she turned 5 years old, every month she collects one shell. By her 10th birthday, how many shells will Martha have collected?}

\end{tcolorbox}
\captionof{figure}{A example for hallucinating}
\label{hallucinate7}

\begin{tcolorbox}[fonttitle=\bfseries]
\normalsize\textbf{Original Solution:}\\
\small{ First, we need to find the total number of houses in all 81 blocks. To do this, we can multiply the number of houses in each block (12) by the number of blocks (81):\\
12 x 81 = 972\\
So there are 972 houses in total. \\
Next, we need to find out how many pieces of junk mail the mailman needs to give to each house. \textcolor{red}{We know that he has to give 4 pieces of junk mail to each house, so we can multiply this by the total number of houses:\\
4 x 972 = 3,888}\\
Therefore, the mailman needs to give 3,888 pieces of junk mail in each block. \textcolor{red}{\ding{55}}}\\
\\
\normalsize\textbf{Original Problem:}
\small{A mailman has to give 4 pieces of junk mail to each house in each of the 81 blocks. If there are 12 houses in each block \textcolor{red}{How many pieces of junk mail should he give in each block?}}\\\\
\normalsize\textbf{Reconstructed Problem:}
\small{There are 81 blocks in a neighborhood, and each block has 12 houses. The mailman needs to deliver 4 pieces of junk mail to each house. \textcolor{red}{How many pieces of junk mail does the mailman need to deliver in total?}}

\end{tcolorbox}
\captionof{figure}{A example for \probthree{}}
\label{misunderstand1}

\subsection{Examples that compare RCoT feedback with human feedback}
\label{app:humanprompt}
\begin{center}
\begin{tcolorbox}[fonttitle=\bfseries, title = {\textcolor{cyan}{Original Problem:} Kelly has 22 nintendo games. How many does she need to buy so that she will have 140 games left?\\
\textcolor{cyan}{Original Solution:} To calculate how many Nintendo games Kelly needs to buy so that she will have 140 games left, we need to subtract the number of games \textcolor{red}{she currently has from the desired total number of games. The desired total number of games is 140 + 22 = 162 games.} Therefore, Kelly needs to buy 162 - 22 = 140 Nintendo games. Therefore, the final answer is 140.\\
\textcolor{cyan}{Reconstructed Problem:} Kelly currently has 22 Nintendo games. She wants to have 140 more games than she currently has. How many Nintendo games does Kelly need to buy to reach her goal?
},before upper = \tcbsubtitle{\hspace{1.2cm}RCoT},
    before lower = \tcbsubtitle{Reasons By Humans}, subtitle style = {
      frame empty,
      fontupper = \sffamily\bfseries,
      colback = cyan!90!black,
      coltext = black!70,
      width = 5cm,
      arc = 8pt,
      rounded corners = east
    }]

\normalsize\textbf{Fine-grained Feedback:}\\
\small{Here are the mistakes and reasons in your answer to the problem.\\\\
You use some wrong candidate conditions:\\
1: She wants to have 140 more games than she currently has.\\
They all can not be deduced from the true condition list.\\
Here are detailed reasons:\\
1: The true problem only provides information about how many games Kelly currently has and she needs to reach a certain number of games. It does not provide any information that Kelly wants to have 140 more games than she currently has. Therefore, the candidate condition of "She wants to have 140 more games than she currently has" cannot be deduced from your thinking.\\\\
The real problem has the conditions:\\
1: Kelly has 22 nintendo games. \\
2: She wants to have 140 games left.\\
You should consider all real conditions in the problem.}\\\\
\normalsize\textbf{Revised Answer:}\\
\small{Thank you for pointing out the mistakes in my previous answer. Here's a corrected response:\\
To calculate how many Nintendo games Kelly needs to buy so that she will have 140 games left, we need to subtract 140 from the total number of games she wants to have. The total number of games she wants to have is the current number of games she has plus the number of games she wants to buy. So, we can set up an equation:\\
22 + x = 140\\
where x is the number of games Kelly needs to buy.\\
Solving for x, we get:\\
x = 140 - 22 = 118\\
Therefore, Kelly needs to buy 118 Nintendo games so that she will have 140 games left.}  \textcolor{darkpastelgreen}{\ding{52}}
\tcblower
\normalsize\textbf{Fine-grained Feedback:}\\
\small{Your answer is wrong. The problem mentions that she will have 140 games left, but it doesn't mean that she desires 140 more games. You misunderstand it.}\\\\
\normalsize\textbf{Revised Answer:}\\
\small{I apologize for the mistake in my previous response. To calculate how many Nintendo games Kelly needs to buy so that she will have 140 games left, we need to subtract 140 from the number of games she currently has. \\
So, if Kelly currently has 22 Nintendo games and wants to have 140 games left, we can calculate the number of games she needs to buy as follows:\\
140 (desired number of games left) - 22 (current number of games) = 118\\
Therefore, Kelly needs to buy 118 Nintendo games to have 140 games left.}  \textcolor{darkpastelgreen}{\ding{52}}

\end{tcolorbox}
\captionof{figure}{This example shows that feedback generated by RCoT sometimes has the same quality as the human-written feedback.}
\label{human_example1}
\end{center}

\begin{center}
\begin{tcolorbox}[fonttitle=\bfseries, title = {\textcolor{cyan}{Original Problem:} There are 11 different books and 17 different movies in the ' crazy silly school ' series. If you read 7 of the books and watched 21 of the movies How many more movies than books have you read?\\ 
\textcolor{cyan}{Original Solution:} 
First, we need to find out how many books and movies are left unread/watched ......
\textcolor{red}{Let's assume that the person has actually watched 17 movies, but some of them were repeats. So, we can say that they have watched 17 unique movies and some of them multiple times.}
Therefore, the person has watched 10 more movies than books they have read.\\
\textcolor{cyan}{Reconstructed Problem:} A person has 11 books and 21 movies. They have read/watched 7 books and 17 movies. How many more movies than books have they watched?},before upper = \tcbsubtitle{\hspace{1.2cm}RCoT},
    before lower = \tcbsubtitle{Reasons By Humans}, subtitle style = {
      frame empty,
      fontupper = \sffamily\bfseries,
      colback = cyan!90!black,
      coltext = black!70,
      width = 5cm,
      arc = 8pt,
      rounded corners = east
    }]

\normalsize\textbf{Fine-grained Feedback:}\\
\small{Here are the mistakes and reasons in your answer to the problem.\\\\
You use some wrong candidate conditions:\\
1: The person has 11 books and 21 movies.\\
They all can not be deduced from the true condition list.\\
Here are detailed reasons:\\
1: The candidate condition states that the person has 11 books and 21 movies, but the condition list only tells us that there are 11 books and 17 movies in the series, and that the person has read 7 of the books and watched 21 of the movies. We cannot deduce that the person has all 11 books just because there are 11 books in the series. Therefore, the candidate condition cannot be deduced from the condition list.\\\\
The real problem has the conditions:\\
1: There are 11 different books and 17 different movies in the 'crazy silly school' series.\\
2: You read 7 of the books and watched 21 of the movies.\\
You should consider all real conditions in the problem.}\\\\
\normalsize\textbf{Revised Solution:}\\
\small{I apologize for the mistakes in my previous answer. Here is the corrected answer:
The true problem has the condition:\\
There are 11 different books and 17 different movies in the 'crazy silly school' series.\\
To find out how many more movies than books you have read ...... \textcolor{red}{Let's assume that you have actually watched 17 movies}, but some of them were repeats. So, we can say that you have watched 17 unique movies and some of them multiple times.\\
Therefore, you have watched 10 more movies than books you have read.}  \textcolor{red}{\ding{55}}
\tcblower
\normalsize\textbf{Fine-grained Feedback:}\\
\small{The books and movies in crazy silly school are not relevant to calculate how many more movies than books you have read. You only need to consider the number of books and movies you have read/watched.}\\\\
\normalsize\textbf{Revised Solution:}\\
\small{You are correct. I apologize for my mistake. To find out how many more movies than books you have read, we simply need to subtract the number of books you have read from the number of movies you have watched:\\
Number of movies watched - Number of books read = 21 - 7 = 14\\
Therefore, you have watched 14 more movies than books you have read.}  \textcolor{darkpastelgreen}{\ding{52}}

\end{tcolorbox}
\captionof{figure}{This example shows RCoT sometimes cannot perform as well as humans.}
\label{human_example2}
\end{center}

\subsection{A concrete example that shows the whole process of RCoT}
\label{app:whole}

\begin{center}
\tcbset{fonttitle=\bfseries}
\begin{tcolorbox}[title={\textcolor{cyan}{\large Original Problem:} Lana had 72 extra nickels. If she put them into stacks with 8 in each stack, how many stacks could she make?\\
\textcolor{cyan}{\large Original Solution: }To calculate how many stacks Lana could make with 72 extra nickels, we need to divide the total number of nickels by the number of nickels in each stack. \textcolor{red}{Since there are 5 nickels in each stack, 72 extra nickels is equal to 72/5 = 14.4 stacks.} However, Lana cannot have a fraction of a stack, so we need to round down to the nearest whole number. Therefore, Lana could make 14 stacks with the 72 extra nickels. Therefore, the final answer is 14. \textcolor{red}{\ding{55}}}, breakable]

\normalsize\textbf{Step1: Problem Reconstruction}\\
\small{\textbf{Input} = [instruction, original solution]\\
\textbf{Output} = [reconstructed problem]}\\\\
\small{\textbf{Instruction:} Give the concrete prompt (problem) that can generate this answer. The problem should contain all basic
and necessary information and correspond to the answer. The problem can only ask for one result.\\
\textbf{Reconstructed Problem:} Lana has 72 extra nickels and \textcolor{blue}{wants to stack them in piles of 5.} How many stacks of nickels can she make?}
\tcblower
\normalsize\textbf{Step2: Fine-grained Comparison}\\
\small{
\textbf{(i)Problem Decomposition:}\\
\small{\textbf{Input} = [instruction, original problem, reconstructed problem]\\
\textbf{Output} = [original condition list, reconstructed condition list]}\\\\
\textbf{Instruction:} Please list the conditions of the problem. There may be multiple conditions......\\
\textbf{Original condition list:}\\
1: Lana had 72 extra nickels.\\
\textcolor{red}{2: She put them into stacks with 8 in each stack.}\\
\textbf{Reconstructed condition list:}\\
1: Lana has 72 extra nickels.\\
\textcolor{blue}{2: She wants to stack them in piles of 5.}\\\\\\
\textbf{(ii)Condition Comparison:}\\
\small{\textbf{Input} = [instruction, sub-condition, condition list]\\
\textbf{Output} = [yes or no, reason]}\\\\
\textbf{Overlooked conditions:}\\
\textbf{Instruction:} Given a candidate condition: \textcolor{red}{"She put them into stacks with 8 in each stack."}\\
Here is a condition list:\\
"1: Lana has 72 extra nickels. 2: She wants to stack them in piles of 5."\\
From a mathematical point of view, can this candidate condition be deduced from the condition list?\\
Please illustrate your reason and answer "yes" or "no".\\\\
\textbf{hallucinated conditions:}\\
\textbf{Instruction:} Given a candidate condition: \textcolor{blue}{"She wants to stack them in piles of 5."}\\
Here is a condition list:\\
"1: Lana had 72 extra nickels. 2: She put them into stacks with 8 in each stack."\\
From a mathematical point of view, can this candidate condition be deduced from the condition list?\\
Please illustrate your reason and answer "yes" or "no".\\\\
\textbf{(iii)Question Comparison:}\\
\small{\textbf{Input} = [instruction, original problem, reconstructed problem]\\
\textbf{Output} = [yes or no, reason]}\\\\
Q1: Lana had 72 extra nickels. If she put them into stacks with 8 in each stack, how many stacks could she make?\\
Q2: Lana has 72 extra nickels and wants to stack them in piles of 5. How many stacks of nickels can she make?\\
\textbf{Instruction:} From a mathematical point of view, are these problems asking the same thing in the end? Please illustrate your reason and answer "yes" or "no".
\tcbline
\normalsize\textbf{Step3: Fine-grained Feedback and Revision}\\
\small{\textbf{Input} = [instruction, original solution]\\
\textbf{Output} = [revised solution]}\\\\
\textbf{Instruction:}\\
Here are the mistakes and reasons in your answer to the problem.\\\\
\textcolor{red}{You have ignored some real conditions:\\
1: She put them into stacks with 8 in each stack.}\\
You should consider all real conditions in the problem.\\
Here are detailed reasons: \\
1: There is no information about whether or not She put them into stacks with 8 in each stack. Therefore, the real condition of ”She put them into stacks with 8 in each stack“ cannot be logically derived or deduced from your thinking.\\\\
\textcolor{blue}{You use some wrong candidate conditions:\\
1: She wants to stack them in piles of 5.}\\
They all can not be deduced from the true condition list.\\
You should consider all real conditions in the problem.\\
Here are detailed reasons:\\
1: The condition list only provides information about the number of extra nickels Lana has and she wants to put 8 nickels in each stack. There is no mention of her preference or intention to stack them in piles of 5.\\\\
\textbf{Revised Solution:}\\
I apologize for my mistake in the previous answer. Here is the corrected answer:\\
To calculate how many stacks Lana could make with 72 extra nickels, we need to divide the total number of nickels by the number of nickels in each stack. \textcolor{darkpastelgreen}{Since there are 8 nickels in each stack, Lana could make 72/8 = 9 stacks with the 72 extra nickels.} Therefore, the final answer is 9.
\textcolor{darkpastelgreen}{\ding{52}}}
\end{tcolorbox}
\captionof{figure}{This is an example showing the whole process of RCoT. \textcolor{red}{Red}: The \edtwo{} condition in the original problem. \textcolor{blue}{Blue}: The \edone{} condition in the original problem. \textcolor{darkpastelgreen}{Green}: The correct solution.}
\label{method_example}
\end{center}

\clearpage
\subsection{An example comparing RCoT and Double-Check}
\label{app:rcotdc}
\begin{figure}[h]
  \centering
 \includegraphics[scale=0.9]{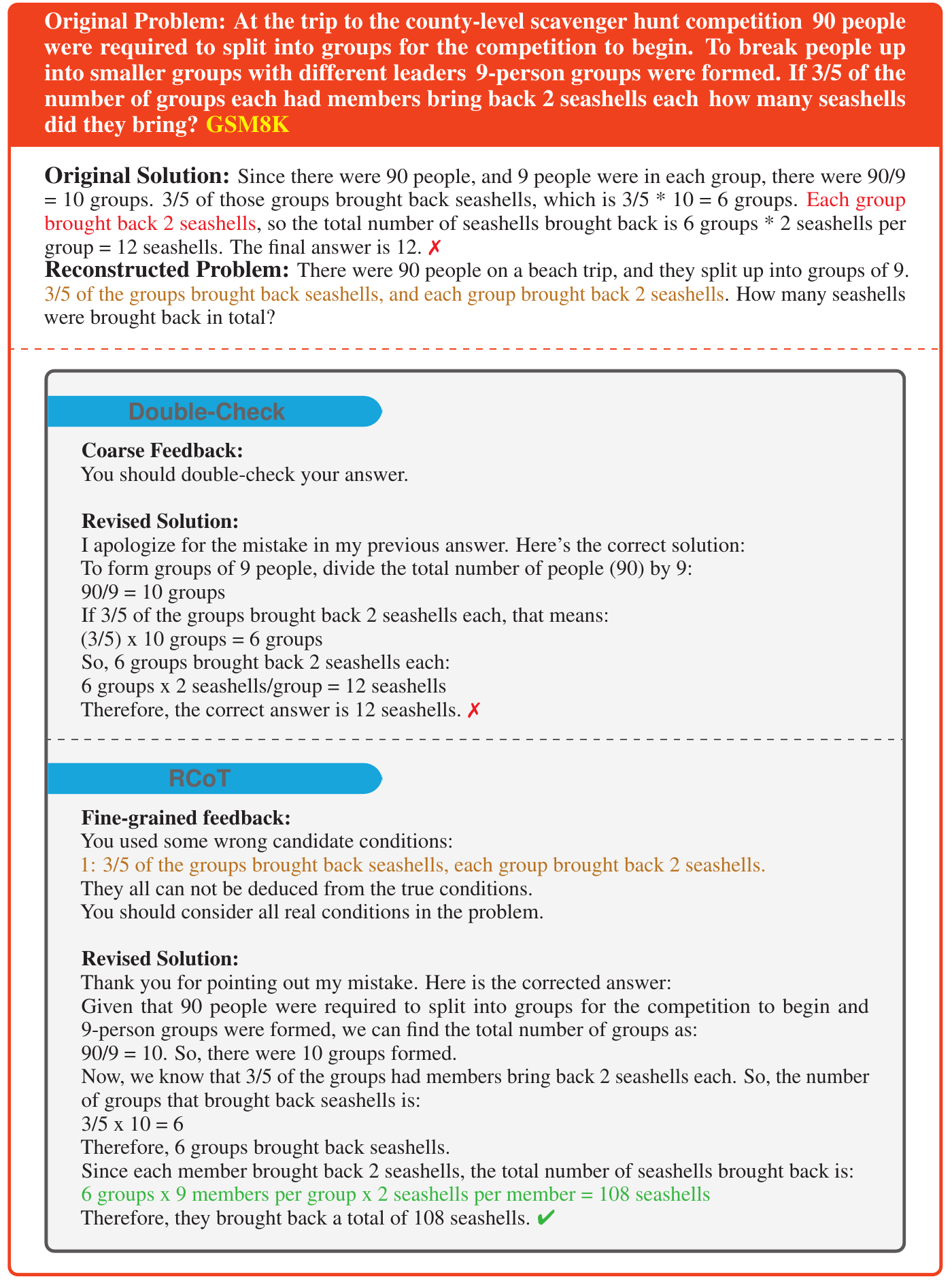}
  \caption{This is an example that Double-Check fails to correct the solution, whereas RCoT succeeded in correcting the solution. \textcolor{darkpastelgreen}{Green}: The correct solution of the problem. \textcolor{red}{Red}: The wrong intermediate step of the original solution. \textcolor{brown}{Brown}: The factual inconsistencies found by RCoT.}
  \label{figure1}
\end{figure}

\clearpage
\subsection{Datasets}
\label{app:dataset}
\begin{table}[h]
\renewcommand{\arraystretch}{1.5}
\caption{\small{Examples of each reasoning task and detailed description of each dataset.}}
\vspace{1mm}
  \label{tab:dataset}
  \centering
  \begin{tabular}{cccp{1cm}p{7cm}}
    \hline
    Dataset &Answer Format&Train &Test &Example
     \\\hline
     GSM8K&Number&7473&1319&{Joseph had 3 times as many notebooks as Martha. Martha decided she needed more notebooks and then bought 5 more for a total of 7 notebooks. How many more than Joseph does she now have?}
      \\\hline
     AQuA&Multiple choice&97467&254&{A man spends 70\% of his income. If his income increases by 20\%, then what will be his new expenditure? Answer Choices: (A) 58.3\% (B) 62.5\% (C) 63.5\% (D) 64.5\% (E) 65.5\%}
      \\\hline
     AddSub&Number&-&395&{Mary is baking a cake . The recipe wants 8 cups of flour . She already put in 2 cups . How many cups does she need to add ?}
      \\\hline
     SVAMP&Number&-&1000&{Bobby ate 28 pieces of candy. Then he ate 42 more. He also ate 63 pieces of chocolate. How many pieces of candy did Bobby eat?} 
      \\\hline
     SingleEq&Number&-&508&{There were 28 bales of hay in the barn. Tim stacked more bales in the barn today. There are now 54 bales of hay in the barn. How many bales did he store in the barn ?}
      \\\hline
     ASDiv&Number&-&2096&{The following week, they decided to go to Lake Huron and Lake Michigan. During their stay there, they caught a total of 30 pikes, 40 sturgeons and 75 herrings. How many fishes did they catch from the two lakes?}
      \\\hline
    Date&Multiple Choices&-&370&{Today is 9/7. Jane is watching NFL 2003. What is the date tomorrow in MM/DD/YYYY? Answer Choices: (A) 08/18/2003 (B) 09/08/1916 (C) 09/13/2003 (D) 09/15/2003 (E) 09/01/2003 (F) 09/08/2003}
      \\\hline
  \end{tabular}
\end{table}

\clearpage
\subsection{An example comparing RCoT and Double-Check}
\label{app:rcotdc}
\begin{figure}[h]
  \centering
 \includegraphics[scale=0.9]{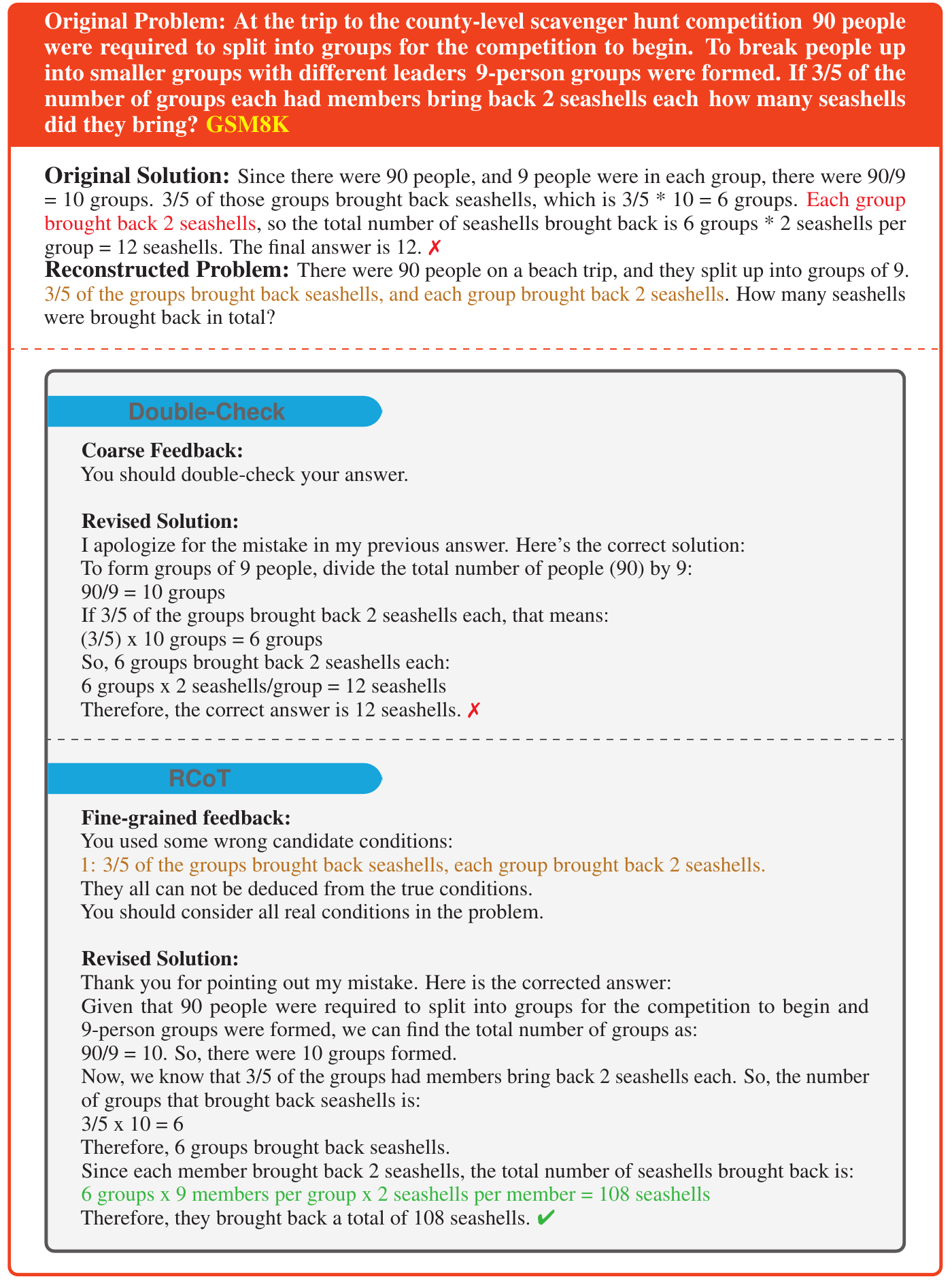}
  \caption{This is an example that Double-Check fails to correct the solution, whereas RCoT succeeded in correcting the solution. \textcolor{darkpastelgreen}{Green}: The correct solution of the problem. \textcolor{red}{Red}: The wrong intermediate step of the original solution. \textcolor{brown}{Brown}: The factual inconsistencies found by RCoT.}
  \label{figure1}
\end{figure}

\subsection{Template}
\begin{figure}[ht]
\centering
\includegraphics[scale=0.9]{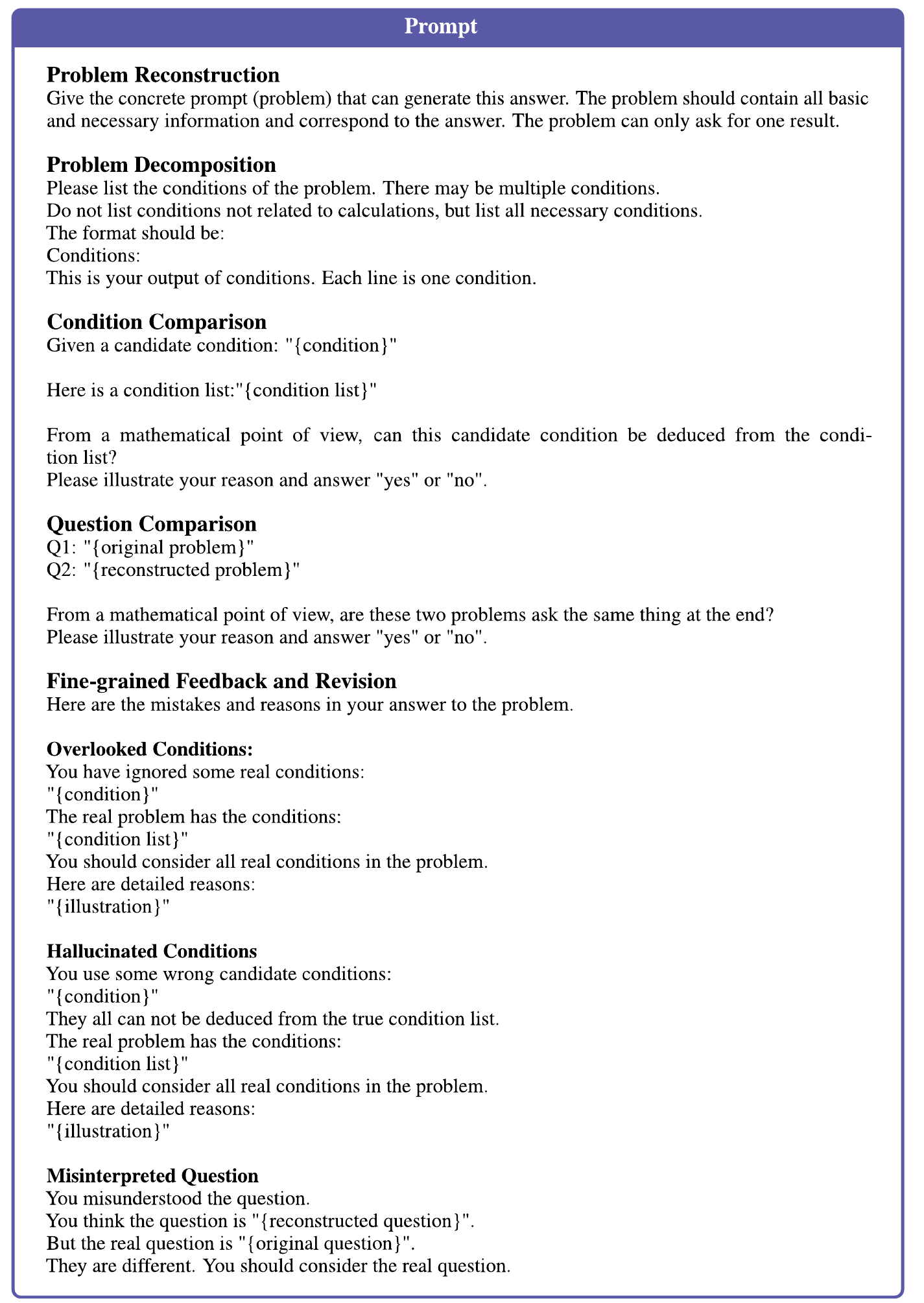}
\caption{All prompts used in RCoT}
\label{prompt_temp}
\end{figure}

Figure \ref{prompt_temp} shows the template prompts of RCoT.

\section{Limitations and Future Work}
\label{app:limit}
RCoT can not detect all possible reasoning errors. For example, it is hard for RCoT to detect computational errors. However, RCoT could be combined with other prompting techniques such as Program-of-Thought \citep{chen2022program}, a method to reduce computational errors through disentangling reasoning and computations. Besides, there is still a significant gap between revising the solutions with RCoT-generated feedback and human feedback, which encourages  further exploration in the generation of fine-grained feedback with higher quality. RCoT requires multiple conversations with LLMs (e.g., ChatGPT in our paper) and may thus slow down the inference speed due to the low bandwidth of API calls. Nevertheless, a locally deployed model may alleviate such a problem. In the future, we will explore other applications of RCoT, such as logical reasoning and symbolic reasoning.

\end{document}